\documentclass[11pt,a4paper]{article}
\usepackage[hyperref]{emnlp-ijcnlp-2019}
\pdfoutput=1
\usepackage{times}
\usepackage{latexsym}
\usepackage{fixltx2e}
\usepackage{textgreek}

\usepackage{url}

\aclfinalcopy % Uncomment this line for the final submission

\newcommand*\samethanks[1][\value{footnote}]{\footnotemark[#1]}

\title{Understanding Roles and Entities: Datasets and Models for Natural Language Inference }

\author{Arindam Mitra\thanks{\quad These authors contributed equally to this work.} \quad  Ishan Shrivastava\samethanks \quad  Chitta Baral\\
Arizona State University \\
{\tt \{amitra7,ishrivas,chitta\}@asu.edu} }

\begin{document}
\maketitle
\begin{abstract}
  We present two new datasets and a novel attention mechanism for Natural Language Inference (NLI). Existing neural NLI models, even though when trained on existing large datasets, do not capture the notion of entity and role well and often end up making mistakes such as ``Peter signed a deal" can be inferred from ``John signed a deal". The two datasets have been developed to mitigate such issues and make the systems better at understanding the notion of ``entities" and ``roles". After training the existing architectures on the new dataset we observe that the existing architectures does not perform well on one of the new benchmark. We then propose a modification to the ``word-to-word" attention function which has been uniformly reused across several popular NLI architectures. The resulting architectures perform as well as their unmodified counterparts on the existing benchmarks and perform significantly well on the new benchmark for ``roles" and ``entities". 
\end{abstract}

\section{Introduction}
Natural language inference (NLI) is the task of determining the truth value of a natural language text, called ``hypothesis'' given another piece of text called ``premise''. The list of possible truth values include \textit{entailment}, \textit{contradiction} and \textit{neutral}. \textit{Entailment} means the hypothesis must be true as the premise is true. \textit{Contradiction} indicates that the hypothesis can never be true if the premise is true. \textit{Neutral} pertains to the scenario where the hypothesis can be both true and false as the premise does not provide enough information. Table \ref{tab:snli} shows an example of each of the three cases. 

\begin{table}[!htb]
    \centering
    \small
    \begin{tabular}{|p{230pt}|}
        \hline
         \textbf{premise:} A soccer game with multiple males playing. \\
         \textbf{hypothesis:} Some men are playing a sport.\\
         \textbf{label:} Entailment.\\\hline
         \textbf{premise:} A black race car starts up in front of a crowd of people. \\
         \textbf{hypothesis:} A man is driving down a lonely road.\\
         \textbf{label:} Contradiction.\\\hline
          \textbf{premise:} A smiling costumed woman is holding an umbrella. \\
         \textbf{hypothesis:} A happy woman in a fairy costume holds an umbrella.\\
         \textbf{label:} Contradiction.\\\hline
    \end{tabular}
    \caption{Example premise-hypothesis pairs from SNLI dataset with human-annotated labels.}
    \label{tab:snli}
    \vspace{-5pt}
\end{table}
Recently several large scale datasets have been produced to advance the state-of-the-art in NLI. One such dataset is SNLI which contains a total of 570k premise-hypothesis pairs. However, several top performing systems on SNLI struggle when they are subjected to examples which require understanding the notion of entity and semantic roles. Table \ref{tab:adv} shows some examples of this kind. 

\begin{table}
    \centering
    \small
    \begin{tabular}{|p{230pt}|}
        \hline
         \textbf{premise:} John went to the kitchen. \\
         \textbf{hypothesis:} Peter went to the kitchen.\\\hline
         \textbf{premise:} Kendall lent Peyton a bicycle. \\
         \textbf{hypothesis:}  Peyton lent Kendall a bicycle.\\\hline
    \end{tabular}
    \caption{Sample premise-hypothesis pairs where existing models trained on SNLI suffers significantly.}
    \label{tab:adv}
    \vspace{-5pt}
\end{table}

The top-performing models on the SNLI benchmark predict \textit{entailment} as the correct label for both the examples in Table \ref{tab:adv} with very high confidence.  For example, the ESIM \cite{DBLP:journals/corr/ChenZLWJ16} model predicts entailment with a confidence of  $82.21\%$ and $96.29\%$ respectively.  

To help the NLI systems to better learn these concepts of entity and semantic roles we present two new datasets. Our contributions are twofold: 1) we show how existing annotated corpus such as VerbNet \cite{Schuler:2005:VBC:1104493}, AMR \cite{banarescu2013abstract} and QA-SRL\cite{DBLP:journals/corr/abs-1805-05377} can be used to automatically create premise-hypothesis pairs that stress on the understanding of entities and roles. 2) We propose a novel neural attention for NLI which combines vector similarity with symbolic similarity to perform significantly better on the new datasets.

%Natural Language Inference (NLI) is the task of deciding whether a natural language sentence is followed or contradicted by another natural language sentence. Recently, several annotated datasets and neural architectures have been developed to address this task. However, the existing architectures trained on the available datasets do not capture the notion of ``entity" and ``roles" well and  often end up making mistakes of the following kinds: ``Peter signed a deal" can be inferred from ``John signed a deal" or ``Peter gave the ball to John" is entailed by ``John gave the ball to Peter". To make the systems better at understanding the notion of `entities" and `role", we create a dataset containing helpful examples. After training the existing architectures on the new dataset we observe that the existing architectures does not perform well on the new benchmark. We then propose a modification to the `word-to-word" attention function which has been uniformly reused across several popular NLI architectures. The resulting architectures perform as well as their unmodified counterparts on the existing benchmarks and perform significantly well on the new benchmark for ``roles" and ``entities". After training the existing architectures on the new dataset we observe that the existing architectures does not yet fully understand the notion of ``entities" and ``roles". 

\section{Dataset Generation}
We create two new datasets. The first one contains examples of \textit{neutral} or \textit{contradiction} labelled \textit{\textit{premise-hypothesis}} pairs where the \textit{hypothesis} is created from the \textit{premise} by replacing its named entities with a different and disjoint set of named entities. This dataset is referred to as \textsc{NER-Changed}. The second one contains examples of \textit{neutral} labelled \textit{\textit{premise-hypothesis}} pairs where the  hypothesis is created by swapping the two different entities from the \textit{premise} which has the same (VerbNet) type but plays different roles. This one is referred to as the \textsc{Role-Swapped}. To help the NLI systems to learn the importance of these modifications, the two datasets also contain \textit{entailment} labelled \textit{\textit{premise-hypothesis}} pairs where the \textit{hypothesis} is exactly same as the \textit{premise}.

\subsection{\textsc{NER-Changed} DataSet}
To create this data set, we utilize the sentences from the bAbI \cite{weston2015towards} and the AMR \cite{banarescu2013abstract} corpus. %With bAbI corpus, we only changed "Name" entities, while with AMR corpus, we changed ``Name'', ``Number'' and also ``Date'' entities to create the ``Neutral'' labelled \textit{premise-hypothesis} pairs.

\subsubsection{Creation of examples using bAbI}
We extract all the $40814$ sentences which contains a single person name and the $4770$ sentences which contain two person names. For all the single name sentences, we replace the name in the sentence with the token \textbf{personX} to create a set of template sentences. For example, the following sentence:
\begin{quote}
  ``Mary moved to the hallway.''
\end{quote}
becomes
\begin{quote}
  ``\textbf{personX} moved to the hallway.''
\end{quote}
This way, we create a total of 398 unique template sentences, each consisting only one name. We then use a list of 15 Gender Neutral names to replace the token \textbf{personX} in all the template sentences. We then make pairs of premise and hypothesis sentences and label the ones with different names as \textit{neutral} and with same name as \textit{entailment}. The template mentioned above, creates the following \textit{\textit{premise-hypothesis}} pair:

\vspace{5pt}
\noindent\fbox{
 
    \parbox{0.98\linewidth}{
        \textbf{Premise} : Kendall moved to the hallway. \newline
        \textbf{Hypothesis} : Peyton moved to the hallway. \newline
        \textbf{Gold Label}: Neutral}
}
\vspace{5pt}

Similarly, we use the two name sentences and the gender neutral names to create more \textit{neutral} labelled \textit{premise-hypothesis} pairs. We ensure that the set of unique template sentences and gender neutral names are disjoint for train, dev, test set.

\subsubsection{Creation of examples using AMR}
Contrary to the bAbI dataset, AMR corpus contains complex and lengthier sentences which provides varity to our dataset. We use the annotation available in the AMR corpus to extract 945 template sentences such that each of them contain at least one mention of a city or a country or a person. Consider the following example with the mention of a city:
\begin{quote}
  ``Teheran defied international pressure by announcing plans to produce more fuel for its nuclear program.''
\end{quote}
Using a list of certain names of cities, countries and persons selected from the AMR corpus we change the names mentioned in the candidate sentences to create the ``Neutral'' labelled \textit{premise-hypothesis} pair. From the example mentioned above, the following pair is generated:

\vspace{10pt}
\noindent\fbox{
 
    \parbox{0.98\linewidth}{
        \textbf{Premise} : \underline{Dublin} defied international pressure by announcing plans to produce more fuel for its nuclear program. \newline
        \textbf{Hypothesis} : \underline{Shanghai} defied international pressure by announcing plans to produce more fuel for its nuclear program.\newline
        \textbf{Gold Label}: Neutral}
}
\vspace{5pt}

We also use the AMR corpus to collect sentences containing ``Numbers'' and ``Dates'' to create \textit{neutral} or \textit{contradiction} labelled \textit{premise-hypothesis} pair. The gold labels in this case is decided manually. The following pair provides an example of this case: 

\vspace{10pt}
\noindent\fbox{
 
    \parbox{0.98\linewidth}{
        \textbf{Premise} : The Tajik State pays \underline{35 dirams} (a few cents) per day for every person in the rehabilitation clinics. \newline
        \textbf{Hypothesis} : The Tajik State pays \underline{62 dirams} (a few cents) per day for every person in the rehabilitation clinics.\newline
        \textbf{Gold Label}: Contradiction}
}
\vspace{5pt}

We also convert a few numbers to their word format and replace them in the sentences to create \textit{premise-hypothesis} pairs. Consider the following example:

\vspace{10pt}
\noindent\fbox{
 
    \parbox{0.98\linewidth}{
        \textbf{Premise} : The Yongbyon plant produces $4$ megawatts. \newline
        \textbf{Hypothesis} : The Yongbyon plant produces five megawatts.\newline
        \textbf{Gold Label}: Contradiction}
}
\vspace{5pt}

\subsection{\textsc{Roles-Switched} DataSet}
The \textsc{Roles-Switched} dataset contains sentences such as ``John rented a bicycle to David'', where two person play two different roles even though they participate in the same event (verb). We use the VerbNet\cite{Schuler:2005:VBC:1104493} lexicon to extract the set of all verbs (events) that take as arguments two same kinds of entities for two different roles. We use this set to extract annotated sentences from VerbNet\cite{Schuler:2005:VBC:1104493} and QA-SRL\cite{DBLP:journals/corr/abs-1805-05377}, which are then used to create sample \textit{premise-hypothesis} pairs. The following two subsections describe the process in detail.

\subsubsection{Creation of dataset using VerbNet}
VerbNet\cite{Schuler:2005:VBC:1104493} provides a list of VerbNet class of verbs and also provides the restrictions defining the types of thematic roles that are allowed as arguments. It also provides a list of member verbs for each class of verbs. For example, consider the VerbNet class for the verb give - ``give-13.1''. The roles it can take are ``Agent'', ``Theme'' and ``Recipient''. It further provides the restrictions as ``Agent'' and ``Recipient'' can only be either an “Animate” or an “Organization” type of entity. 

We use this information provided by VerbNet\cite{Schuler:2005:VBC:1104493} to shortlist $47$ VerbNet classes (verbs) that accepts the same kind of entities for different roles. ``give-13.1'' is one such class as the two different roles for it, ``Agent'' and ``Recipient'' accepts the same kind of entities, namely ``Animate'' or ``Organization''. We take the member verbs from each of the shortlisted VerbNet classes to compute the set of all $646$ ``interesting'' verbs.  We then extract the annotated sentences from VerbNet to finally create the template sentences for the data set creation. 

Consider the following sentence from VerbNet which contains the verb ``lent" which is a member verb of the VerbNet class ``give-13.1''.
\begin{quote}
  ``They lent me a bicycle.''
\end{quote}
We use such sentences and associated annotations to create template sentences such as:
\begin{quote}
``\textbf{PersonX} lent \textbf{PersonY} a bicycle.''
\end{quote}

Note that VerbNet provides example sentence for each VerbNet classes not for individual member verbs and sometimes the example sentence might not contain the required \textbf{PersonX} and \textbf{PersonY} slot. Thus, using this technique, we obtain  a total of $89$ unique template sentences from VerbNet. For all such template sentences, we use gender neutral names to create the neutral labelled role-swapped \textit{premise-hypothesis} pairs, as shown below:

\vspace{10pt}
\noindent\fbox{
 
    \parbox{0.98\linewidth}{
        \textbf{Premise} : Kendall lent Peyton a bicycle. \newline
        \textbf{Hypothesis} : Peyton lent Kendall a bicycle.\newline
        \textbf{Gold Label}: neutral}
}
\vspace{5pt}

%We take special care in considering the verbs that although take same kind of entities as arguments for different roles, but switching them among the roles doesn't necessarily generate a different meaning. Consider the verb ``accompanied". Switching the arguments for the two roles for this verb will not change the meaning of the sentence. For such verbs, we created ``entailment" labelled \textit{premise-hypothesis} pairs. 

\subsubsection{Creation of dataset using QA-SRL}
In the QA-SRL\cite{DBLP:journals/corr/abs-1805-05377} dataset, roles are represented as questions. Thus we go through the list of questions from the QA-SRL\cite{DBLP:journals/corr/abs-1805-05377} dataset to  map the questions into their corresponding VerbNet role. We consider only those QA-SRL\cite{DBLP:journals/corr/abs-1805-05377} sentences which contains both the role-defining  questions of a verb in their annotation and where each of the entity associated with those two roles (the answer to the questions) is either a singular or a  plural noun, or a singular or a plural proper noun. We then swap those two entities to create a \textit{neutral} labelled \textit{premise-hypothesis} pair. The following pair shows an example:

\vspace{10pt}
\noindent\fbox{
 
    \parbox{0.98\linewidth}{
        \textbf{Premise} : Many kinds of power plant have been used to drive propellers.\newline
        \textbf{Hypothesis} : Propellers have been used to drive many kinds of power plant.\newline
        \textbf{Gold Label}: neutral}
}
\vspace{5pt}

% Consider the question ``What does something become''. We extract all the annotated sentences from the QA-SRL data set that contains this question and consider all the answers to this question to compute replacement candidates for creating the \textit{premise-hypothesis} pairs. An example of a sentence which contains the question ``What does something become'' is:

% \begin{quote}
% ``Mercury is emitted as a gas , but as it cools , it becomes a droplet.'' \newline
% \end{quote}
% where the answer to the question us ``Mercury''. For choosing the best replacement candidate from the list of replacement candidates, we compute the glove vector similarity and chose the least similar option as the best replacement candidate. This gave us the following neutral labelled \textit{premise-hypothesis} pair for the example mentioned above:
% \vspace{10pt}
% \noindent\fbox{
 
%     \parbox{0.98\linewidth}{
%         \textbf{Premise} : Mercury is emitted as a gas , but as it cools , it becomes a droplet.\newline
%         \textbf{Hypothesis} : Mercury is emitted as a gas , but as it cools , it becomes a thorium-234.\newline
%         \textbf{Gold Label}: neutral}
% }
% \vspace{5pt}

\begin{table*}[!t]
\centering
\begin{tabular}{|@{}p{1.5cm}|@{}p{0.6cm}|@{}p{1cm}|@{}p{1cm}|p{1cm}|p{1cm}|p{1cm}|p{1cm}|p{1cm}|p{1cm}|p{1cm}|@{}p{1cm}|  }
 \hline
 \multicolumn{2}{|c|}{\textbf{Data Sets} } & \multicolumn{2}{|c|}{\textbf{DecAtt} } & \multicolumn{2}{|c|}{\textbf{ESIM} }& \multicolumn{2}{|c|}{\textbf{Lambda DecAtt} } & \multicolumn{2}{|c|}{\textbf{Lambda ESIM} }& \multicolumn{2}{|c|}{\textbf{BERT} }    \\
 \hline 
     Train &Test &Train\newline Acc &Test\newline Acc &Train\newline Acc &Test\newline Acc
     &Train\newline Acc &Test\newline Acc
     &Train\newline Acc &Test\newline Acc
     &Train\newline Acc &Test\newline Acc\\
  \hline
 \footnotesize{SNLI} &\footnotesize{NC} &84.58\% &\textbf{59.34}\%	&89.78\% &33.59\%	&85.1\% &46.48\%	&90.10\% &33.08\%	&91.59\% &8.37\%
 \\ 
 \hline
 \footnotesize{SNLI + NC} &\footnotesize{NC}	&85.58\% &88.43\%	&89.42\% &51.96\%	&85.8\% &\textbf{96.14}\%	&89.72\% &92.61\%	 &90.97\% &80.55\%
 \\
  \hline
\footnotesize{SNLI + NC}  &\footnotesize{SNLI}	 &85.58\% &84.12\%	 &89.42\% &87.27\%	 &85.8\% &84.41\%	 &89.72\%  &87.19\%	 &90.97\% &\textbf{89.17}\%
	\\
  \hline				
\footnotesize{SNLI}   &\footnotesize{RS}	 &84.58\%  &54.62\%	 &89.78\%  &53.96\%	 &85.1\%  &\textbf{54.72}\%	 &90.10\%  &53.96\%	 &91.59\%  &20.81\%
\\
  \hline
\footnotesize{SNLI + RS}  &\footnotesize{RS}	 &85.25\%  &75.12\%	 &89.93\%  &87.33\%	 &84.24\%  &77.38\%	 &90.3\%  &\textbf{90.29}\%	 &90.84\%  &72.15\%
\\
  \hline
\footnotesize{SNLI + RS}  &\footnotesize{SNLI} 	 &85.25\% &85.20\%	 &89.93\% &88.21\%	 &84.24\% &84.56\%	 &90.3\% &87.74\%	 &90.84\% &\textbf{88.88}\%
	\\
  \hline				
\footnotesize{SNLI + RS + NC}  &\footnotesize{NC}	 &86.49\% &92.05\%	 &89.69\% &53.46\%	 &86.4\% &\textbf{97.24}\%	 &90.7\% &95.88\%	 &90.72\% &80.55\%
\\
  \hline
\footnotesize{SNLI + RS + NC}  &\footnotesize{SNLI}	 &86.49\% &84.72\%	 &89.69\% &87.09\%	 &86.4\% &84.26\%	 &90.7\% &87.81\%	 &90.72\% &\textbf{89.09}\%
\\
  \hline
\footnotesize{SNLI + RS + NC}  &\footnotesize{RS}	 &86.49\% &76.09\%	 &89.69\% &88.86\%	 &86.4\%	&77.85\% &90.7\% &\textbf{90.76}\%	 &90.72\% &68.50\% \\
 \hline
 
\end{tabular}

\caption{\label{tab:widgets}Table shows the train and test set accuracy for all the experiments. Here, NC refers to \textsc{NER-Changed} dataset and RS refers to the \textsc{Role-Switched} dataset. Each row of this table represents an experiment. The first two columns of each row represents the train set and the test set used for that experiment. Rest of the columns show the train and the test accuracy (Acc) in percentages for all the five models. In our experiments, we have used the \textit{bert-large-uncased} model.}
\end{table*}

\section{Model}
In this section we describe the existing attention mechanism of the DecAtt \cite{DBLP:journals/corr/ParikhT0U16} and the ESIM \cite{DBLP:journals/corr/ChenZLWJ16} model. We then describe the proposed modification which helps to perform better on the \textsc{NER Changed} dataset.

Let $a$ be the premise and $b$ be the hypothesis with length \textit{l\textsubscript{a}} and \textit{l\textsubscript{b}} such that a = (\textit{a\textsubscript{1},a\textsubscript{2},...,a\textsubscript{l\textsubscript{a}}}) and b = (\textit{b\textsubscript{1},b\textsubscript{2},...,b\textsubscript{l\textsubscript{b}}}) where each \textit{a\textsubscript{i}} and \textit{b\textsubscript{j}} $\in$ $\mathcal{R}$\textsuperscript{d} is a word vector embedding of dimensions \textit{d}.

Both DecAtt and the ESIM model first transforms the original sequence $a$ and $b$ to another sequence $\bar a $ = ($\bar a_1,...,\bar a_{l_a}$) and $\bar b$ = ($\bar b_1, ...,\bar b_{l_b}$) of same length to learn task-specific word embeddings. It then computes a non normalized attention between each pair of words using dot product as shown in equation \ref{attentiion_orig}.

\begin{equation}
\label{attentiion_orig}
e\textsubscript{ij} = (\bar a\textsubscript{i}) \textsuperscript{\textit{T}}\bar b\textsubscript{j}
\end{equation}

Since the initial word embeddings for similar named entities such as ``john'' and ``peter'' are very similar, the normalized attention scores between \textsc{NER-Changed} sentence pairs such as `` Kendall moved to the hallway.'' and ``Peyton moved to the hallway.''  forms a diagonal matrix which normally occurs when premise is exactly same as hypothesis. As a result, the systems end up prediction \textit{entailment} for this kind of premise-hypothesis pairs. To deal with this issue, we introduce symbolic similarity into the attention mechanism.  The attentions scores are then computed as follows:
\begin{equation}
e'\textsubscript{ij} = \lambda \textsubscript{ij} e\textsubscript{ij} + (1-\lambda \textsubscript{ij})sym\textsubscript{ij}
\end{equation}

Here, \textit{sym\textsubscript{ij}} represents the symbolic similarity which is assigned 0 if the string representing a\textsubscript{i} is not ``equal'' to the string representing b\textsubscript{j}. If the two string matches, then a weight \textit{w} which is a hyper-parameter, is assigned.  $\lambda_{ij} \in [0,1]$  is a learnable parameter which decides how much weight should be given to vector similarity and how much weight to the symbolic similarity (\textit{sym\textsubscript{ij}}) while calculating the new unnormalized attention weights \textit{e'\textsubscript{ij}}. $\lambda$\textsubscript{ij} is computed using equation \ref{lamda}. We will refer to this feed-forward neural network as the lambda layer.

\begin{equation}
\label{lamda}
\lambda_{ij} = 1-LeakyReLU(1-LeakyReLU(W_\lambda x^\lambda_{ij}))
\end{equation}

Here, W\textsubscript{$\lambda$} is learned from data with respect to the NLI task and $x^\lambda_{ij}$ is the input to the lambda layer which is a $16$ dimensional sparse feature vector and encodes the NER (Named Entity Recognition) information for the pair of words in the two sentences. We group the NER information into 4 categories namely `Name", ``Numeric", ``Date" and ``Other".   We use Spacy and Stanford NER tagger to obtain the NER category of a word. Let $v^{ner}_i$ and $v^{ner}_j$ be two vectors in $\{0,1\}^4$ which encode the one-hot representation of the NER category, then $x^\lambda_{ij} [k_1*4+k_2]$ = $v^{ner}_i[k_1] * v^{ner}_j[k_2]$ where $k_1$ and $k_2$  $\in \{0,1,2,3\}$.

\section{Related Works}
Many large labelled NLI datasets have been released so far. \newcite{DBLP:journals/corr/BowmanAPM15} develop the first large labelled NLI dataset containing $570k$ premise-hypothesis pairs. They show sample image captions to crowd-workers and the label (entailment, contradiction and neutral) and ask workers to write down a hypothesis for each of those three scenarios. As a result they obtain a high agreement entailment dataset known as Stanford Natural Language Inference (SNLI). Since premises in SNLI contains only image captions it might contain sentences of limited genres. MultiNLI \cite{DBLP:journals/corr/WilliamsNB17} have been developed to address this issue. Unlike SNLI and MultiNLI, \cite{scitail} and \cite{DBLP:journals/corr/abs-1809-02922} considers multiple-choice question-answering as an NLI task to create the SciTail \cite{scitail} and QNLI \cite{DBLP:journals/corr/abs-1809-02922} datasets respectively.  Recent datasets like PAWS \cite{2019arXiv190401130Z} which is a paraphrase identification dataset also helps to advance the field of NLI. \newcite{DBLP:journals/corr/abs-1805-02266} creates a NLI test set which shows the inability of the current state of the art systems to accurately perform inference requiring lexical and world knowledge.

Since the release of such large data sets, many advanced deep learning architectures have been developed \cite{DBLP:journals/corr/BowmanGRGMP16,2015arXiv151106361V,DBLP:journals/corr/MouMLXZYJ15,DBLP:journals/corr/LiuSLW16,DBLP:journals/corr/MunkhdalaiY16,2015arXiv150906664R,DBLP:journals/corr/WangJ15b,DBLP:journals/corr/ChengDL16,DBLP:journals/corr/ParikhT0U16,2016arXiv160704492M,sha-etal-2016-reading,DBLP:journals/corr/PariaADCP16,DBLP:journals/corr/ChenZLWJ16,scitail,DBLP:journals/corr/abs-1810-04805,DBLP:journals/corr/abs-1901-11504}. Although many of these deep learning models achieve close to human level performance on SNLI and MultiNLI datasets, these models can be easily deceived by simple adversarial examples. \newcite{DBLP:journals/corr/abs-1805-04680} shows how simple linguistic variations such as negation or re-ordering of words deceives the DecAtt Model. \newcite{DBLP:journals/corr/abs-1803-02324} goes on to show that this failure is attributed to the bias created as a result of crowd sourcing. They observe that crowd sourcing generates hypothesis that contain certain patterns that could help a classifier learn without the need to observe the premise at all.
\section{Experiments}

We split the \textsc{NER-Changed} and \textsc{Role-Switched} dataset in train/dev/test sets each containing respectively 85.7K/4.4k/4.2k and 10.4/1.2k/1.2k premise-hypothesis pairs, which is then used to evaluate the performance of a total of five models. This includes three existing models, namely DecAtt \cite{DBLP:journals/corr/ParikhT0U16}, ESIM \cite{DBLP:journals/corr/ChenZLWJ16} and BERT \cite{DBLP:journals/corr/abs-1810-04805} and  two new models namely Lambda DecAtt (ours) and Lambda ESIM (ours). The results are shown in Table \ref{tab:widgets}.

We observe that if the models are trained on the SNLI train set alone, they perform poorly on the \textsc{NER-Changed} and \textsc{Role-Switched} test set . This could be attributed to the absence of similar examples in the SNLI dataset. After exposing the new datasets at train time along with the SNLI training dataset, DecAtt and BERT shows significant improvement where the ESIM model continues to struggle in the \textsc{NER-Changed} test set. Our Lambda DecAtt and Lambda ESIM models however significantly outperform the remaining models and achieves as well as or better accuracy than its unmodified counterparts DecAtt and ESIM on the SNLI test set.

\section{Conclusion}
We have shown how the existing annotated meaning representation datasets can be used to create NLI datasets which stress on the understanding of entities and roles. Furthermore, we show that popular existing models when trained on existing datasets hardly understand the notion of entities and roles. We have proposed a new attention mechanism for natural language inference. As experiments suggest, the new attention function significantly helps to capture the notion of entities and roles. Furthermore, the performance does not drop on the existing testbeds when the new attention mechanism is used, which shows the generality of the proposed attention mechanism.

\bibliography{emnlp-ijcnlp-2019}
\bibliographystyle{acl_natbib}
\end{document}